\documentclass[11pt,a4paper]{article}
\usepackage[hyperref]{rocling2026}
\usepackage{times}
\usepackage{latexsym}

\usepackage{graphicx}
\usepackage{amsmath}
\usepackage{array}
\usepackage{booktabs}
\usepackage{mdframed}
\usepackage{float}
\usepackage{capt-of}
\usepackage{multirow}
\usepackage{fontawesome5}
\usepackage{academicons}

\usepackage{microtype}
\usepackage{enumitem}

\roclingfinalcopy 

\title{\textsc{ConvoDrift}: A Multi-Turn Conversational Dataset for Modeling Stylistic Tone Evolution}

\author{
Vihindi Kotalawala$^\spadesuit$,
Pamoda Dilranga$^\diamondsuit$,
Gayani Thoradeniya$^\clubsuit$,
Prasan Yapa$^\heartsuit$ \\
$^\spadesuit$Informatics Institute of Technology, Sri Lanka \\
$^\diamondsuit$Independant Researcher, Sri Lanka \\
$^\clubsuit$National Health and Medical Research Council, Australia \\
$^\heartsuit$Luxembourg Centre for Systems Biomedicine (LCSB), University of Luxembourg \\
\texttt{vihindi.20220672@iit.ac.lk, pamoda9@gmail.com} \\
\texttt{gayani.thoradeniya@nhmrc.gov.au, prasan.yapa@uni.lu}
}

\date{}

\begin{document}
\maketitle

\begin{abstract}
The evolution of linguistic style in conversations is an underexplored issue in NLP. Most style-control datasets focus on sentences or assume a static style throughout, missing the dynamic shifts that occur as user preferences change during interactions.
We introduce \textsc{ConvoDrift}, a dataset designed to model progressive stylistic conversational tone drift under fixed semantic intent. It is built on 15{,}727 shared multi-turn conversational structures for adaptation and persona-conditioned alignment methods. It consists of six prompt--response pairs per conversation, each with the annotation of style drift and style direction labels. These pairs cover a range of communication genres. We further derive a complementary pairwise dataset by pairing semantically equivalent but stylistically distinct responses and annotating persona-conditioned preferences using five distinct style communication personas, enabling the controlled study of personalisation and pluralistic alignment in language tone. In addition to dataset construction, we conduct a comprehensive evaluation involving human validation, LLM-as-judge assessment, and automatic lexical and semantic evaluations. Across seven Likert criteria annotated by three human annotators, the average \emph{Krippendorff’s $\alpha$} is 0.88, and our lexical and semantic analyses show that drift events induce lexical changes while preserving semantic similarity.

\end{abstract}


\section{Introduction}

Conversational interactions involving language models are inherently dynamic: users often adjust linguistic style and tone across multiple turns as goals, preferences, and stylistic expectations evolve. However, existing Natural Language Processing (NLP) datasets and benchmarks provide limited support for modelling this kind of progressive, turn-by-turn stylistic change. In particular, most style-control and style-transfer benchmarks operate at the sentence level \cite{rao-tetreault-2018-dear} and explicitly ignore conversational context, making them ill-suited for studying stylistic evolution over multi-turn dialogue. \cite{roy-etal-2023-conversation}. This highlights the lack of datasets that capture how tone progressively drifts during a conversation while preserving the context and maintaining the original semantic intent.

We introduce \textsc{ConvoDrift}\footnote{
Dataset is available at \href{https://github.com/Vihindi/_CONVODRIFT_/tree/main}{\faGithub\ Github} and
\href{https://huggingface.co/collections/Vihindi-K/convodrift} {\raisebox{-0.30em}{\includegraphics[height=1.5em]{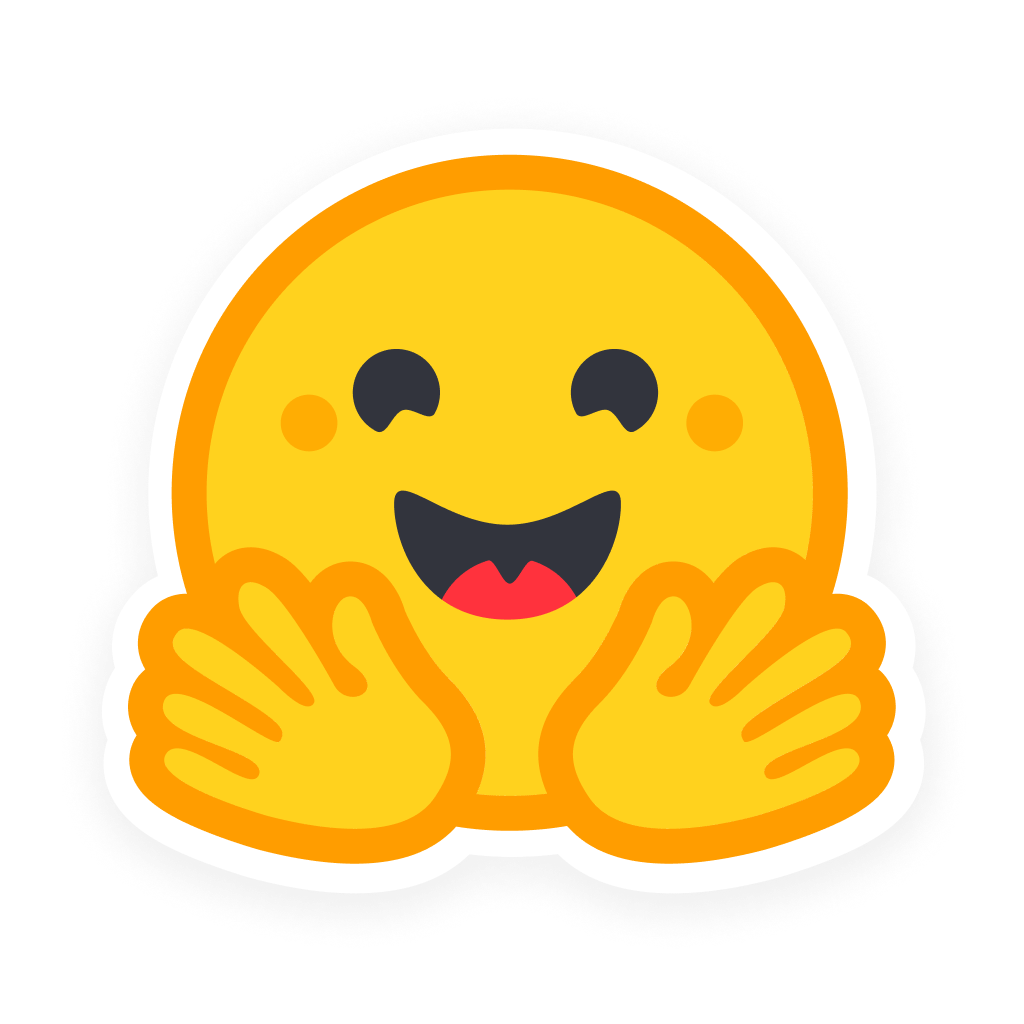}} HuggingFace}.
}, a flexible dataset built around a shared conversational benchmark. This corpus facilitates the study of progressive stylistic tone drift in multi-turn dialogue and the derivation of style and tone-relevant preference data suitable for personalised alignment via Reinforcement Learning From Human Feedback (RLHF), addressing existing limitations in the field. By explicitly separating semantic content from stylistic modification, \textsc{ConvoDrift} enables controlled analysis of tone variation across turns without confounding content changes, a property that is difficult to achieve with existing dialogue corpora~\cite{cheng-etal-2020-contextual}. The same conversational trajectories can be reused for multiple purposes, including modelling stylistic dynamics or serving as preference supervision, and can be evaluated under different preference profiles through persona-conditioned tone drift. This design naturally supports pluralistic alignment, in which multiple stylistic preferences are treated as valid rather than collapsed into a single reward signal, addressing concerns highlighted in recent alignment research~\cite{poddar2024personalizing}. To further demonstrate the practical utility of the dataset, we conducted a downstream preference modelling task using Variational Preference Learning (VPL)~\cite{poddar2024personalizing}. The results show that \textsc{ConvoDrift} can support persona-conditioned preference learning and is suitable for training preference models that capture diverse stylistic preferences.

In summary, \textsc{ConvoDrift} contributes: (1) Explicit supervision for progressive tone drift in multi-turn dialogue; (2) A clear separation between semantic content and stylistic modification; (3) Support for studying persona-conditioned and pluralistic stylistic preferences within a unified conversational setting.

\section{Related Work}
\subsection{Stylistic Tone Variation and Drift in Text}
Research on stylistic control in text has historically focused on style transfer, in which a single input sentence is rewritten to exhibit a target stylistic attribute while preserving its semantic content. Foundational datasets such as the Grammarly Yahoo Answers Formality Corpus (\texttt{GYAFC})~\cite{rao-tetreault-2018-dear} enable supervised informal–formal rewriting and have become standard benchmarks for formality transfer. The existing research has explored stylistic attributes, including politeness~\cite{danescu-niculescu-mizil-etal-2013-computational,niu-bansal-2018-polite}, sentiment \cite{fu2018style} , professionalism~\cite{ficler-goldberg-2017-controlling}, and controlled generation~\cite{hu2017toward} without parallel data~\cite{shen2017style}. Despite covering a range of stylistic attributes, these approaches predominantly operate at the sentence level, framing stylistic control as a single-turn rewriting task applied to isolated utterances. In these formulations, stylistic transformation is treated as a context-independent operation, where each input is processed without access to conversational history or discourse structure~\cite{ficler-goldberg-2017-controlling,hu2017toward}. As a result, such methods do not capture stylistic dynamics across a conversation, where tone may evolve gradually in response to prior turns or communicative goals.

More recent work has begun to incorporate dialogue context into stylistic modelling. \citet{roy-etal-2023-conversation} propose a few-shot conversational style transfer framework that rewrites entire dialogues conditioned on exemplar conversations. Nevertheless, this line of work focuses on global style transformation rather than planned, turn-by-turn stylistic drift, leaving fine-grained conversational tone evolution underexplored.

Although conversational style transfer incorporates dialogue context, it still provides limited supervision for \emph{when} stylistic changes occur and \emph{how} tone evolves progressively across turns. Modelling such tone evolution also introduces a personalisation dimension, as different users may exhibit and prefer different stylistic behaviours throughout an interaction. This observation naturally connects stylistic drift modelling to research on personalised and persona-conditioned dialogue systems, which explicitly account for user-specific conversational behaviours.

This connection between stylistic variation and user-specific behaviour motivates consideration of persona-based dialogue research. While such research captures user identity and preferences at the level of content and conversational consistency it offers limited mechanisms for modelling how stylistic preferences (e.g., tone and writing style) are learned, compared, or optimised during training, despite their central role in personalisation~\cite{zhang2024personalization}. Addressing this limitation requires preference-based learning frameworks that explicitly model human judgements, motivating the growing role of preference data and RLHF.

\subsection{Personalised Dialogues and Alignment}

Personalisation in dialogue systems has traditionally been approached through persona conditioning, where responses are generated based on user-specific attributes or profiles~\cite{zhang-etal-2018-personalizing}. \citet{li-etal-2016-persona} demonstrated that conditioning on textual persona descriptions improves speaker consistency and engagement in open-domain dialogue. Subsequent research extended this paradigm by incorporating demographic and contextual attributes such as age, gender, and emotion into dialogue systems~\cite{zhao-etal-2025-personalens}, influencing response content and style.
While prior work shows that personas shape stylistic behaviour~\cite{joshi2017personalization,zhao2025teaching,lee-etal-2025-picpersona}, these approaches typically treat style as a static property of a user or conversation, rather than as a signal that evolves across interaction turns.

At the same time, surveys on personalised language models highlight tone and writing style as key personalisation factors, alongside relevance and accuracy~\cite{zhang2024personalization}. Effective personalisation, therefore, requires models to adapt to users’ stylistic preferences across different contexts, rather than producing an average response. However, most existing preference datasets do not explicitly label style or capture how stylistic alignment changes over multi-turn conversations. This limitation motivates the development of conversational datasets that jointly model style, persona, and preference dynamics over time.

RLHF has become the dominant paradigm for aligning language models with human preferences, typically using pairwise comparisons aggregated into a single reward signal~\citep{ouyang2022training,bai2022training,ziegler2019fine}. While recent work explores personalisation and pluralistic alignment by modelling heterogeneous or user-conditioned preferences~\cite{poddar2024personalizing,park2024rlhf,cheng2023everyone}, existing RLHF datasets primarily focus on global qualities such as helpfulness, safety, and correctness. Stylistic factors, particularly tone and writing style, are rarely annotated explicitly, and preference signals are generally treated as static rather than context-dependent.

Taken together, prior work on personalised dialogue and preference-based alignment provides mechanisms for modelling \emph{who} the speaker is and \emph{what} they prefer, but offers limited support for capturing \emph{how} stylistic expression evolves over the course of a conversation. Most datasets either abstract away conversational context or collapse preferences into a single reward signal, limiting their ability to capture fine-grained, progressive stylistic changes. These limitations highlight shortcomings in existing resources for studying dynamic tone adaptation and stylistic alignment in multi-turn conversations.

\section{Methodology}
\begin{figure*}[t]
    \centering
    \includegraphics[width=0.9\textwidth]{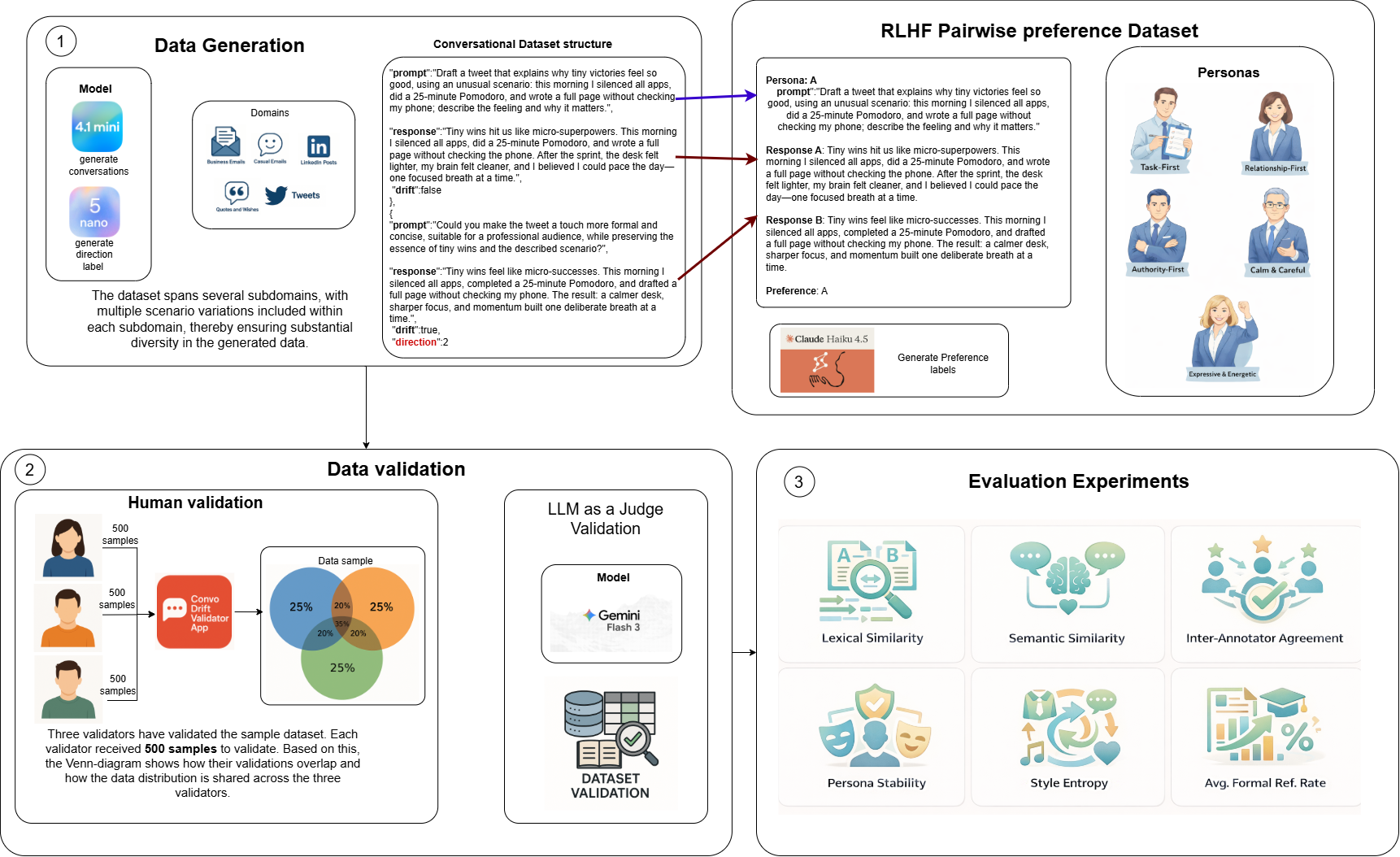}
    \caption{
    Overview of the ConvoDrift data construction and evaluation pipeline.
    \textbf{$^1$ Data Generation} illustrates the creation of structured multi-turn conversations across diverse domains and scenarios using selected language models, followed by the derivation of stylistically contrasting responses for RLHF pairwise preference data with persona-conditioned profiling. 
    \textbf{$^2$ Data Validation} shows the quality assurance process, combining human validation via a custom annotation interface and LLM-as-a-judge validation. 
    \textbf{$^3$ Evaluation Experiments} summarises the downstream evaluation protocols applied to the resulting datasets.
    }
    \label{fig:framework}
\end{figure*}
\textsc{ConvoDrift} is a multi-stage data generation and validation dataset designed to support both the analysis of multi-turn stylistic tone drift and persona-conditioned preference learning for personalised RLHF. Accordingly, the dataset consists of two core stages, each producing a distinct but tightly coupled dataset. (1) \textsc{ConvoDrift}--Conversation: A multi-turn conversational dataset that captures progressive stylistic tone drift under fixed semantic intent. (2) \textsc{ConvoDrift}--RLHF: A persona-conditioned pairwise preference dataset constructed from conversational trajectories to enable personalised and pluralistic alignment.

As shown in Figure~\ref{fig:framework}, both datasets reuse the same multi-turn prompts and responses. The RLHF subset restructures responses from these conversations into style-contrasting preference pairs. This unified design ensures consistency between conversational structure and preference supervision, enabling \textsc{ConvoDrift} to serve as both a diagnostic benchmark for tone drift and a resource for studying personalised and pluralistic alignment \cite{poddar2024personalizing}.
To provide empirical evidence of the dataset’s effectiveness, we conducted a downstream preference modelling task using VPL with representations of GPT-2 and Llama 3.1 8B, achieving accuracies of 0.579 and 0.65, respectively, with further details provided in Appendix~\ref{app:vpl_downstream}.

\subsection{Conversational Dataset Creation}
This section describes how \textsc{ConvoDrift} was constructed to capture progressive conversational tone changes while preserving semantic intent across multiple turns.
\subsubsection{Model Selection and Generation Setup}
\label{sec:model_selection}

To select appropriate models for data generation, we conducted a preliminary comparison across 12 Large Language Models (LLMs) using an identical set of conversational prompts. Models were evaluated based on output quality, adherence to stylistic instructions, latency, and estimated token-level cost under a fixed input--output budget.
Based on this comparison, we selected \texttt{GPT-4.1-mini} as the primary model for conversational response generation due to its consistent output quality and reliable stylistic control, and \texttt{GPT-5-nano} for generating explicit drift, direction labels due to its higher quality and lower latency outputs.
Using this setup, we generated 15{,}727 multi-turn conversations, each containing six prompt--response pairs, resulting in a total of 94{,}362 response turns. This has been discussed in detail in Appendix~\ref{app:Dataset additional details}.

\subsubsection{Prompt Design and Quality Control}

\textsc{ConvoDrift} generates conversations across multiple genres as shown in Table~\ref{tab:genre_counts} with distinct stylistic norms, spanning both professional and informal contexts. Each genre is paired with diverse communicative scenarios, and prompts enforce semantic preservation while allowing controlled stylistic variation. Generated conversations are then filtered using a rule-based quality pipeline to ensure structural consistency, including a fixed conversation length, valid drift annotations, and the presence of multiple stylistic drift events.

\begin{table}[t]
\centering
\small
\resizebox{0.6\columnwidth}{!}{
\begin{tabular}{lr}
\hline
\textbf{Communication genre} & \textbf{Count} \\
\hline
Business email & 2{,}997 \\
Casual email   & 3{,}246 \\
Quotes/Wishes  & 3{,}340 \\
LinkedIn post  & 3{,}150 \\
Tweet          & 2{,}994 \\
\bottomrule
\end{tabular}}
\caption{Conversation counts by communication genre in the \textsc{ConvoDrift} dataset.}
\label{tab:genre_counts}
\end{table}

\subsubsection{Multi-turn Conversational Structure}
\label{sec:conversation_structure}

Each conversation is structured as a sequence of six prompt--response pairs:
{\small
\begin{equation}
C = \{(p_t, r_t)\}_{t=1}^{6}.
\end{equation}
}
where \(p_t\) denotes the user prompt at turn \(t\) and \(r_t\) the corresponding response generated conditionally on the preceding conversational context. Figure~\ref{fig:example_conversation} provides an illustrative example of this six-turn structure. The first prompt reflects the user’s original request, denoted as T1 is always annotated as \texttt{Drift=False}. Subsequent turns refine the previous response based on the user’s prompt; we set \texttt{Drift=True} only when the prompt explicitly requests a style/tone change.

Importantly, semantic content is preserved across turns, enabling the dataset to capture drift trajectories rather than isolated stylistic rewrites~\cite{roy-etal-2023-conversation}.

\begin{figure}[t]
    \centering
    \setlength{\fboxsep}{6pt}
    \fbox{%
    \begin{minipage}{0.95\columnwidth}
    \tiny
    \textbf{Example conversation (Tweet domain; scenario: share a thought about learning through challenges).}\vspace{2pt}

    \textbf{T1 (Drift=False)}\\
    \textbf{User:} I want to share a tweet about how recovery should be viewed as an essential, ongoing part of resilience, not just an end goal.\\
    \textbf{Assistant:} Recovery isn't the finish line---it's a crucial part of the journey. Every step of healing, no matter how small, builds resilience. Embrace the process and honor your progress. \#Resilience \#Healing\\[2pt]

    \textbf{T2 (Drift=True; Direction=2)}\\
    \textbf{User:} Rewrite it in a more formal and precise tone, emphasizing that recovery is a gradual process.\\
    \textbf{Assistant:} Recovery is a gradual process rather than a fixed endpoint. Each stage contributes to resilience and supports sustained personal growth.\\[2pt]

    \textbf{T3 (Drift=False)}\\
    \textbf{User:} Make it shorter and more direct for a busy feed.\\
    \textbf{Assistant:} Recovery isn't the finish line---it's every brave step forward. Healing builds resilience, one moment at a time. \#Resilience \#Healing\\[2pt]

    \textbf{T4 (Drift=True; Direction=1)}\\
    \textbf{User:} Rewrite with a more casual, approachable tone.\\
    \textbf{Assistant:} Hey, recovery's not just about getting to the end---it's every little step you take to heal. Even when it's slow, you're building strength. Keep going. \#Resilience\\[2pt]

    \textbf{T5 (Drift=True; Direction=0)}\\
    \textbf{User:} Include a nature metaphor to symbolise recovery and resilience.\\
    \textbf{Assistant:} Recovery is like a tree growing through the seasons---sometimes slow, sometimes tough, but always reaching for the light. Keep reaching. \#Resilience \#NatureHeals\\[2pt]

    \textbf{T6 (Drift=False)}\\
    \textbf{User:} Keep the tree-growth metaphor and condense the message into two clear sentences.\\
    \textbf{Assistant:} Recovery parallels a tree's growth through seasons---steady and resilient. Each phase of healing strengthens the foundation for future growth.
    \end{minipage}%
    }
    \caption{Example \textsc{ConvoDrift} conversation illustrating progressive tone edits under fixed semantic intent.}
    \label{fig:example_conversation}
    
\end{figure}

\subsubsection{Drift, Direction Labels, and Metadata}
\label{sec:drift_direction}

Each prompt--response pair is annotated with two complementary labels, which are shown in the example conversation in Figure~\ref{fig:example_conversation}.
\begin{itemize} 
    \item A \textit{binary drift label} (\texttt{Drift}), indicating whether the user prompt requests a stylistic/tone change; and
    \item A \textit{direction label} (\texttt{Direction}), indicating the direction of the stylistic shift relative to the previous response.
\end{itemize}

The direction label is defined as:
{\small
\[
\texttt{direction} =
\left\{
\begin{array}{ll}
0, & \textit{no stylistic change} \\
1, & \textit{casual / expressive tone} \\
2, & \textit{formal / precise tone}
\end{array}
\right.
\]
}

The drift label is primarily driven by the user prompt, whereas direction is inferred from the stylistic properties of the generated response. For example, in Figure~\ref{fig:example_conversation}, T4 is marked \texttt{Drift=True} because the user asks for a ``more casual" rewrite, and the corresponding response is labelled with \texttt{Direction=1} to reflect a shift toward a more casual/expressive style. These two labels are evaluated independently to avoid conflating intended stylistic instructions with their realised linguistic effects. Main statistics about the dataset has been discussed in Appendix~\ref{app:statistics}.

\subsection{Conversion to RLHF Pairwise Data}
\label{sec:rlhf_conversion}
This section explains how conversational trajectories are transformed into semantically aligned but stylistically contrasting response pairs for persona-conditioned preference learning.

\subsubsection{Pairwise Preference Construction}
\label{sec:pairwise}

Each multi-turn conversation in \textsc{ConvoDrift} consists of a sequence of prompts that progressively request stylistic refinements to a response, producing a trajectory of responses with shared semantic intent but differing stylistic realisations. 

To derive preference supervision suitable for RLHF, we construct pairwise preference instances by selecting a prompt $p$ from the conversation and pairing two responses associated with it. Specifically, the initial response generated for the task is treated as one candidate ($y_a$), while a later response from the same conversation corresponding to an explicit stylistic drift with a known direction (e.g., casual or formal) is selected as the contrasting candidate ($y_b$). The process of constructing the RLHF pairwise dataset section is shown in Figure~\ref{fig:framework}.

Response pairs may be drawn from adjacent or non-adjacent turns, depending on the desired degree of stylistic contrast. By construction, both responses are semantically equivalent, and preference distinctions arise solely from stylistic variation rather than content differences. Each preference instance is represented as a tuple $(p, y_a, y_b)$, following the standard pairwise preference-learning formulation used in RLHF~\cite{christiano2017deep,ouyang2022training}.

\begin{table}[t]
\centering
\small
\setlength{\tabcolsep}{5pt}
\renewcommand{\arraystretch}{1.15}
\resizebox{0.8\linewidth}{!}{
\begin{tabular}{l|p{0.60\columnwidth}}
\hline
\textbf{Persona} & \textbf{Description} \\
\hline
Task-First &
Focuses on efficiency, clarity, and structured reasoning, prioritising actionable outcomes with minimal unnecessary elaboration. \\\hline
Relationship-First &
Emphasises warmth, empathy, and cooperative language to support positive and sustainable interpersonal interactions. \\\hline
Authority-First &
Communicates with confidence and decisiveness, favouring clear direction, assertive tone, and leadership-oriented language. \\\hline
Calm \& Careful &
Uses measured, precise language to minimise risk, ensure safety, and promote stable and cautious decision-making. \\\hline
Expressive \& Energetic &
Engages with enthusiasm and vivid language, favouring expressive phrasing that creates excitement and momentum. \\
\hline
\end{tabular}}
\caption{Persona profiles used for persona-conditioned preference labelling. Each persona encodes a distinct communication focus that guides stylistic preference judgements.}
\label{tab:personas}
\end{table}

\subsubsection{Persona-based Preference Learning}
\label{sec:persona}

\textsc{ConvoDrift} supports persona-conditioned preference learning by leveraging the explicit stylistic annotations available in the underlying conversational dataset. Since each response is labelled with drift and direction tags (e.g., toward casual/expressive or formal/precise), stylistic preferences can be systematically grouped and evaluated without relying on implicit or noisy preference inference.

As an experimental instantiation, we define five communication personas, each encoding a consistent stylistic bias shown in Table~\ref{tab:personas}. These personas are not intended to model user identity, but rather to represent distinct and interpretable evaluation perspectives over stylistic variation.

For each example, which includes a prompt and two responses, each persona independently chooses the response that best matches its communication style. This produces separate preference labels for each persona. By keeping these persona-specific labels instead of combining them into a single decision, the dataset preserves differences in stylistic preferences. This design allows us to study personalised and pluralistic alignment, in which multiple communication styles are treated as valid at the same time~\cite{cheng2023everyone,poddar2024personalizing}.

\section{Evaluation and Experiments}
\subsection{Conversational Dataset}
This section evaluates the reliability and quality of the conversational dataset through human validation, LLM-as-a-judge assessment, and automatic semantic and lexical analyses.

\subsubsection{Human Evaluation}
\label{sec:human_eval}

We conducted a human evaluation to assess the correctness of tone drift annotations and the overall quality of multi-turn conversations in \textsc{ConvoDrift}. Given the cost and time of manual validation, three independent annotators evaluated a stratified subset of the dataset using a sampling strategy designed to balance coverage with reliable agreement estimation. 

Since human validation is particularly valuable but limited, we used a controlled-overlap design to maximise the number of human-validated conversations while still enabling agreement analysis: 175 conversations were annotated by all annotators (gold set), 300 conversations were annotated by two annotators (100 conversations per annotator pair), and 375 conversations were annotated by a single annotator (125 unique conversations per annotator). This results in 1{,}500 conversation-level ratings in total (500 per annotator), covering 850 unique conversations. The communication genre balance was maintained across all subsets, ensuring each genre was represented in similar proportions. Inter-annotator agreement metrics were computed exclusively on the gold set, while the remaining annotated data was used for qualitative analysis and downstream evaluation.


Annotators evaluated each conversation on seven ordinal criteria using a five-point Likert scale. We measure agreement on these ordinal ratings using Krippendorff’s $\alpha$ for ordinal data, which supports multiple annotators~\cite{krippendorff2011computing}, and report the resulting scores in Table~\ref{tab:kripp_alpha}.

In addition to the scalar ratings, annotators were instructed to correct drift flags and tone direction labels whenever inconsistencies were identified. Drift is annotated as a binary value indicating whether a tone change is requested in the prompt, while direction labels specify the direction of tone change (no change, formal to casual, or casual to formal). For validating these drift and direction labels on the common subset, inter-annotator agreement was Drift $\alpha$: \emph{ 0.7628} and Direction $\alpha$: \emph{0.7772}. Overall, the agreement results indicate substantial reliability for studying conversational stylistic drift and personalised alignment.

\subsubsection{LLM-as-a-Judge Evaluation}

In addition to human validation, we employed an LLM-as-a-judge framework~\cite{zheng2023judging}. After human evaluation, the remaining portion of the dataset was evaluated using \texttt{Gemini 3 Flash} as an automatic judge. The LLM was prompted to assign drift labels, direction labels, and conversation-level quality ratings using the same annotation guidelines followed by human annotators.
To assess the reliability of the LLM judge, its outputs were compared against a gold-standard human reference constructed via majority voting over human annotations. Agreement for the drift and direction labels was measured using Cohen's $\kappa$.
The LLM demonstrates strong agreement with human validation for tone drift detection, achieving Drift of \emph{0.888}, and substantial agreement for tone direction labelling with Direction of \emph{0.930}~\cite{cohen1960coefficient}. These results indicate that the LLM reliably identifies when tone changes occur and correctly captures their directional shift, while aligning with human annotator decisions.

\begin{table}[t]
\centering
\small
\setlength{\tabcolsep}{5pt}
\renewcommand{\arraystretch}{1.12}
\begin{tabular}{p{0.56\columnwidth}|c|c}
\hline
\textbf{Criterion} & $\boldsymbol{\alpha}$ & \textbf{95\% CI} \\
\hline
Tone consistency and drift annotation accuracy &  \multirow{2}{*}{0.763} & \multirow{2}{*}{[0.732, 0.794]} \\\hline
Tone direction label accuracy &  0.777 & [0.751, 0.803] \\\hline
Drift distribution realism & 0.891 & [0.853, 0.926] \\\hline
Fluency and clarity & 0.946 & [0.916, 0.973] \\\hline
Cross-turn coherence & 0.964 & [0.939, 0.985] \\\hline
Prompt-following accuracy & 0.900 & [0.863, 0.936] \\\hline
Domain-specific naturalness & 0.916 & [0.883, 0.947] \\
\hline 
\end{tabular}
\caption{Inter-annotator agreement on the common subset measured using Krippendorff's alpha for ordinal data. Confidence intervals correspond to 95\% bootstrap intervals.}
\label{tab:kripp_alpha}
\end{table}

\subsubsection{Automatic Evaluation}

\vspace{1em}
\noindent
\textbf{Semantic Similarity:}
We evaluate semantic preservation across refinement turns using sentence-level semantic similarity computed with \texttt{SBERT} embeddings~\cite{reimers-gurevych-2019-sentence}. We report three complementary measures: (i) \textit{adjacent-turn similarity}, which compares consecutive turns and captures local semantic continuity; (ii) \textit{anchor similarity}, which compares each turn against the initial response and captures global meaning preservation; and (iii) \textit{end-to-start similarity}, which compares the final turn of a conversation with its initial response to measure start-to-end semantic retention.

As shown in Table~\ref{tab:semantic_turn_overall}, both adjacent-turn and anchor similarity remain high. Adjacent-turn similarity has a mean of 0.899, and anchor similarity has a mean of 0.890, indicating that responses preserve semantic intent both locally and globally as stylistic refinements progress. Additionally, end-to-start similarity is also high, with a mean of 0.852, indicating that meaning is largely retained from the first to the final turn.

Unlike lexical similarity, we do not stratify semantic similarity by drift labels: drift in \textsc{ConvoDrift} is intended to change style, but regardless of whether a prompt requests drift or not, the underlying meaning of responses should be preserved throughout the conversation. Therefore, semantic similarity is evaluated using these three turn-level views of meaning preservation.

We further analyse semantic preservation across communication genres, as depicted in Table~\ref{tab:semantic_domain_end}, which presents domain-wise end-to-start similarity statistics.

\vspace{1em}
\noindent
\textbf{Lexical Similarity: }
While semantic similarity remains high across turns, lexical overlap can change substantially when tone edits are introduced. We evaluate lexical variation across consecutive turns to quantify these surface-level changes. Lexical similarity is computed using \texttt{ROUGE-L} (longest common subsequence overlap), \texttt{Jaccard-1} (unigram overlap), and \texttt{Jaccard-2} (bigram overlap).

Genre-wise lexical similarity within non-drift regions (Within; Drift=False) and at drift boundaries (Boundary; Drift=True) in turns within the conversations is summarised in Table~\ref{tab:lexical_by_genre}. Across all metrics and genres, within transitions consistently exhibit higher overlap than boundary transitions. Overall, the drops in \texttt{ROUGE-L}, \texttt{Jaccard-1}, and \texttt{Jaccard-2} at boundary transitions indicate that tone drift is accompanied by meaningful lexical rephrasing, although the underlying semantic intent is preserved.

\begin{table}[t]
\centering
\small
\setlength{\tabcolsep}{4pt}
\renewcommand{\arraystretch}{1.12}
\begin{tabular}{lcccc}
\hline
\textbf{Similarity Type} & \textbf{n} & \textbf{Mean} & \textbf{Median} & \textbf{Min} \\
\hline
Adjacent (\texttt{SBERT}) & 2500 & 0.8993 & 0.9207 & 0.3287 \\
Anchor (\texttt{SBERT}) & 3000 & 0.8903 & 0.9034 & 0.2382 \\
End-to-Start (\texttt{SBERT}) & 492 & 0.8519 & 0.8715 & 0.4423 \\
\hline
\end{tabular}
\caption{\label{tab:semantic_turn_overall} Overall turn-level semantic similarity across refinement turns.}
\end{table}

\begin{table}[t]
\centering

\small
\setlength{\tabcolsep}{4pt}
\renewcommand{\arraystretch}{1.12}
\begin{tabular}{lcccc}
\hline
\textbf{Genre} & \textbf{\#Convs} & \textbf{Mean} & \textbf{Median} & \textbf{Min} \\
\hline
Business email & 93  & 0.9123 & 0.9255 & 0.7594 \\
Casual email   & 100 & 0.8551 & 0.8679 & 0.6223 \\
LinkedIn post  & 99  & 0.8842 & 0.8919 & 0.7380 \\
Quotes/Wishes  & 100 & 0.7739 & 0.7971 & 0.4423 \\
Tweet          & 100 & 0.8385 & 0.8580 & 0.5517 \\
\hline
\end{tabular}
\caption{\label{tab:semantic_domain_end} Domain-wise end-to-start semantic similarity measured using \texttt{SBERT} across different communication genres.}

\end{table}

\begin{table*}[t]
\centering
\setlength{\tabcolsep}{8pt}
\resizebox{0.7\linewidth}{!}{
\begin{tabular}{l|l|r|r|r|r}
\hline
\textbf{Genre} & \textbf{Type} & \textbf{n} & \textbf{ROUGE-L} & \textbf{Jaccard-1} & \textbf{Jaccard-2} \\
\hline
\multirow{2}{*}{Business email} & Within (Drift=False)   & 6{,}242  & 0.606 & 0.543 & 0.374 \\
 & Boundary (Drift=True) & 8{,}742  & 0.538 & 0.469 & 0.281 \\ \hline
\multirow{2}{*}{Casual email} & Within (Drift=False)   & 7{,}031  & 0.684 & 0.639 & 0.501 \\
 & Boundary (Drift=True) & 9{,}199  & 0.492 & 0.457 & 0.267 \\ \hline
\multirow{2}{*}{LinkedIn post} & Within (Drift=False)   & 6{,}823  & 0.686 & 0.624 & 0.508 \\
 & Boundary (Drift=True) & 8{,}927  & 0.393 & 0.342 & 0.162 \\ \hline
\multirow{2}{*}{Quotes/Wishes}  & Within (Drift=False)   & 5{,}757  & 0.492 & 0.422 & 0.255 \\
 & Boundary (Drift=True) & 10{,}943 & 0.340 & 0.282 & 0.109 \\ \hline
\multirow{2}{*}{Tweet} & Within (Drift=False)   & 5{,}886  & 0.572 & 0.487 & 0.301 \\
 & Boundary (Drift=True) & 9{,}084  & 0.386 & 0.314 & 0.128 \\
\hline
\multirow{2}{*}{Overall} & Within (Drift=False)   & 31{,}739 & 0.614 & 0.550 & 0.396 \\
 & Boundary (Drift=True) & 46{,}895 & 0.426 & 0.369 & 0.186 \\
\hline
\end{tabular}}
\caption{Genre-wise lexical similarity between consecutive turns in conversations within non-drift regions (Within; Drift=False) and at drift boundaries (Boundary; Drift=True), reported with \texttt{ROUGE-L} and unigram/bigram \texttt{Jaccard} ($n$ transitions).}
\label{tab:lexical_by_genre}
\end{table*}

\subsection{Persona-conditioned RLHF Dataset}

This section evaluates the quality of the pairwise preference labels and their ability to capture consistent yet distinct persona-conditioned communication preferences.
\subsubsection{Experimental study} 
We evaluate candidate LLM judges using stability, persona sensitivity, style entropy, and formal preference rate to select a reliable model for persona-conditioned labelling. During the construction of this pairwise dataset, we removed data points with semantic similarity below 0.6 to keep the dataset semantically similar but stylistically distinct.

\vspace{1em}
\noindent
\textbf{Stability Consistency:} Stability measures how consistently an LLM returns the same preference label when asked multiple times under identical conditions. For each sample and persona, we query the model five times and report stability as the fraction of runs that match the majority label (e.g., 5/5 indicates perfect stability).
High stability indicates that the model’s outputs are not dominated by randomness. 

\vspace{1em}
\noindent
\textbf{Persona Sensitivity:} Persona sensitivity measures how strongly an LLM differentiates between personas by quantifying how often personas disagree on the preferred response (formal vs.\ casual) for the same sample.
For each sample annotated by $K$ personas, we compute sensitivity using the \emph{average pairwise disagreement rate} across persona pairs:

{\small
\begin{equation}
\mathrm{Sensitivity} = \frac{1}{\binom{K}{2}} \sum_{i<j} \mathbf{I}(s_i \neq s_j),
\end{equation}
}

where $s_i$ and $s_j$ denote the preference labels inferred for personas $i$ and $j$.
We report sensitivity by averaging this disagreement score across samples.
Low values indicate that most personas agree on the same response, while high values indicate substantial disagreement across personas.

\vspace{1em}
\noindent
\textbf{Style Entropy:} Style entropy measures how consistent or flexible a persona’s preferences are across samples, based on how balanced its selections are between the two styles (formal vs.\ casual).
For each persona, we compute \emph{Shannon entropy} \cite{shannon1948mathematical} (in bits) based on how often it chooses formal vs.\ casual responses:

{\small
\begin{equation}
\mathrm{style\,entropy} 
= - \textstyle\sum_{s \in \{\mathrm{formal}, \mathrm{casual}\}} p(s)\log_2 p(s),
\end{equation}
}

where $p(s)$ denotes the fraction of times the persona selects style $s$ across the evaluation set.
Low entropy indicates that a persona almost always prefers the same style, reflecting highly consistent preferences, while high entropy indicates that a persona selects both styles at more similar rates, reflecting more context-dependent preferences.

\vspace{1em}
\noindent
\textbf{Average Formal Preference Rate:} The average formal preference rate summarises the model’s overall tendency to prefer formal responses across all comparisons, independent of persona.
Low values indicate a global bias toward casual selections, while high values indicate a global bias toward formal selections.
A value close to 0.5 indicates that, overall, the model selects formal and casual at similar rates.

Considering the results in Table~\ref{tab:llm_selection}, \texttt{Claude Haiku 4.5} provides the most suitable balance among the evaluated criteria. It combines very high stability with the strongest persona sensitivity and relatively low global stylistic bias, while maintaining sufficient contextual flexibility.

Based on this analysis, \texttt{Claude Haiku 4.5} is selected as the LLM for generating persona-conditioned preference labels in the final dataset.

\begin{table}[t]
\centering
\small
\setlength{\tabcolsep}{1pt}
\renewcommand{\arraystretch}{1.12}
\begin{tabular}{|l|c|c|c|c|}
\hline
\multirow{2}{*}{\textbf{Model}} & \multirow{2}{*}{\textbf{Stability}} & \multirow{2}{*}{\textbf{Sensitivity}} & \multicolumn{2}{c|}{\textbf{Average}} \\ \cline{4-5}
 & &  & \textbf{Entropy} & \textbf{Formal Rate} \\ \hline
Claude Haiku 4.5 & {0.990} & {0.533} & 0.853 & 0.542 \\ \hline
Gemini 2.5  flash     & 0.942 & 0.504 & 0.868 & 0.581 \\ \hline
GPT-5 Mini       & 0.932 & 0.483 & {0.886} & 0.585 \\ \hline
GPT-5            & 0.956 & 0.442 & 0.884 & 0.604 \\
\hline
\end{tabular}
\caption{Comparison of candidate LLMs for persona-conditioned preference labelling.}
\label{tab:llm_selection}
\end{table}

\subsubsection{Relation to Agreement Analysis}
\label{subsec:agreement_relation}

Inter-persona agreement was assessed using Cohen's $\kappa$ to examine whether the personas make similar or contrasting stylistic preference judgments. The plot shows two broad preference patterns. Personas A, C, and D each have two scores above the $\kappa=0$ line, indicating positive agreement with one another, while Personas B and E have one positive score representing their agreement with each other. The remaining scores fall below zero, showing disagreement between these two groups. Although some personas share similar preferences, differences in the distribution and magnitude of their agreement scores indicate that each persona retains distinct evaluative characteristics. Further details, including the exact pairwise agreement values, are provided in Appendix~\ref{app:pairwise_agreement}.

\begin{figure}[t]
    \centering
    \includegraphics[width=0.9\linewidth]
    {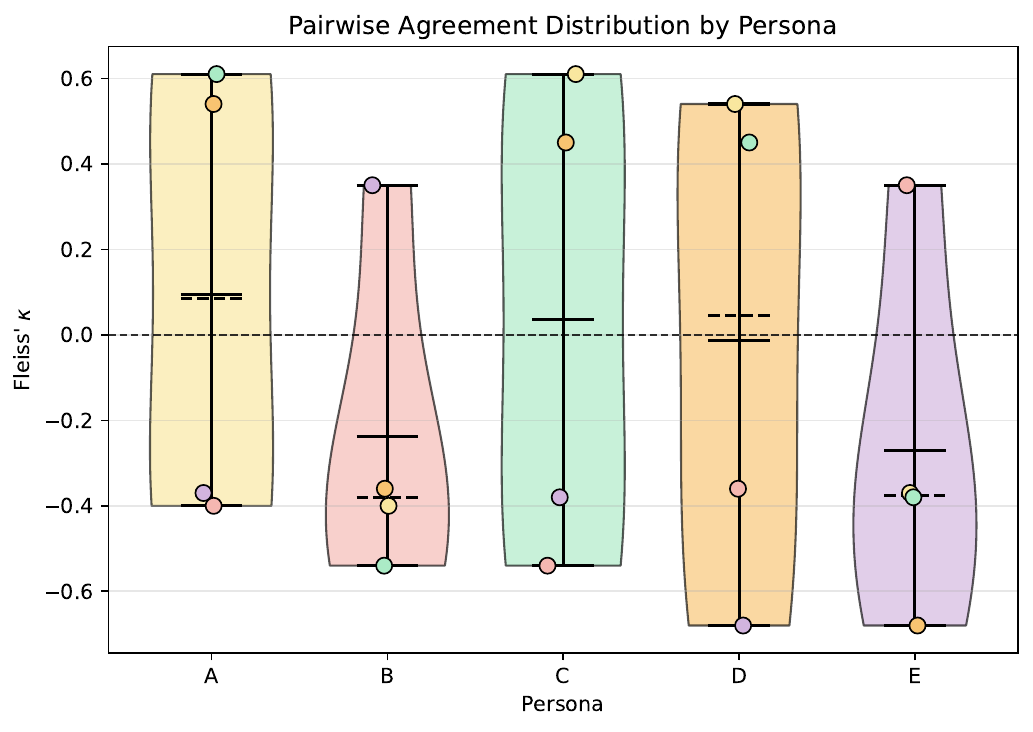 }
    \caption{Distribution of pairwise agreement scores for each persona using Cohen's $\kappa$.}
    \label{fig:persona_agreement_violin}
\end{figure}

\section{Conclusion}

In this paper, we introduced \textsc{ConvoDrift}, a benchmark for modelling progressive stylistic tone drift in multi-turn conversations. By utilising a shared conversational structure, we explicitly separate semantic intent from stylistic variation, enabling the derivation of both multi-turn drift trajectories and persona-conditioned preference datasets for pluralistic alignment. Comprehensive validation via human assessment, LLM-as-a-judge, and lexical metrics confirms that this preserves semantic meaning while capturing structured stylistic dynamics.

\section{Limitations and Future Directions}
There are several avenues remain for future exploration to address the constraints of our current approach. First, as a synthetically generated dataset designed to emulate human–LLM conversations, it lacks the variability and spontaneity inherent in real-world organic data; thus, future work may include real human conversations to capture this realism. Second, we currently measure tone variations only between ``formal" and ``casual" styles, so future research can explore more complex emotions and subtle style shifts. Third, all conversations are constrained to a fixed length of six turns, limiting the analysis of longer-range stylistic dependencies and naturally varying dialogue lengths. Fourth, as the dataset is predominantly synthetically generated and LLM-annotated, it may inherit model-specific stylistic patterns and annotation biases that do not fully reflect naturally occurring human–LLM interactions. Finally, persona conditioning is restricted to five predefined personas; expanding this space to include a broader range of user preferences would further strengthen the dataset’s relevance for personalisation and pluralistic alignment research.



\bibliography{rocling2026}
\bibliographystyle{acl_natbib}

\appendix

\section{Downstream Validation with VPL}
\label{app:vpl_downstream}

To examine the effectiveness of the proposed dataset beyond direct label annotation, we conducted a downstream validation experiment using Variational Preference Learning (VPL) \cite{poddar2024personalizing}. The purpose of this task was to show the effectiveness of the proposed drift and direction labels in providing useful learning signals for preference modelling, and to assess whether the dataset captures meaningful persona-specific preference patterns. VPL is well-suited for this analysis because it represents preference learning as a latent-variable problem, where the latent variable $z$ captures hidden user or annotator context and enables the reward model to predict preferences conditioned on that context, rather than averaging all preferences into a single reward function.
\begin{figure}[H]
    \centering
    \includegraphics[width=1.0\linewidth]{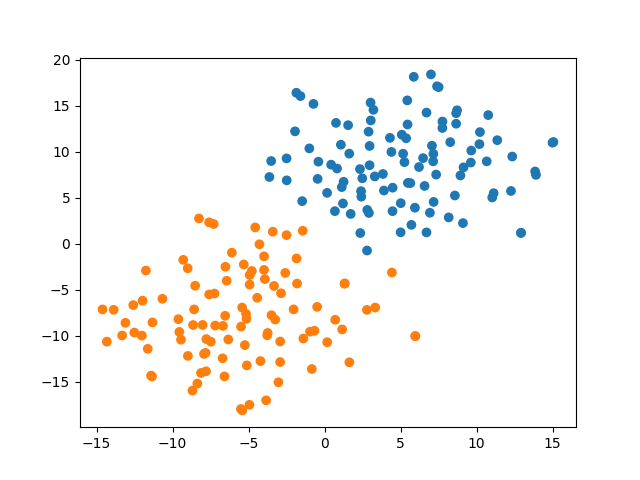}
    \caption{t-SNE visualisation of the latent representations learned by the Llama 3.1 8B-based VPL model.}
    \label{fig:placeholder}
\end{figure}

\begin{figure}[H]
    \centering
    \includegraphics[width=1.0\linewidth]{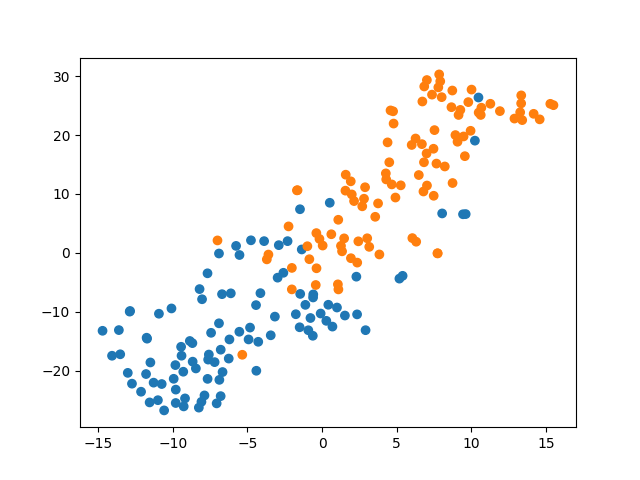}
    \caption{t-SNE visualisation of the latent representations learned by the GPT-2-based VPL model. }
    \label{fig:placeholder}
\end{figure}
Following this setup, we used the prompt together with response A and response B from our pairwise dataset, treating Persona A and Persona B as two distinct preference contexts. The model was trained to predict the preferred response under the corresponding persona-conditioned setting. We evaluated the VPL-based model using two representation backbones: GPT-2 and Llama 3.1 8B. The GPT-2-based model achieved an accuracy of 0.579, while the model using Llama 3.1 8B representations achieved a higher accuracy of 0.65. This indicates that the proposed labels provide predictive signal for distinguishing persona-dependent preferences. The stronger performance of Llama 3.1 8B is likely due to its larger parameter capacity and richer representation quality, which allow it to capture more fine-grained stylistic and persona-conditioned differences.

We further visualised the learned latent representations using t-SNE plots. The resulting visualisations show clearer clustering for the Llama 3.1 8B-based model, with two distinct clusters corresponding to Persona A and Persona B. This separation suggests that the model learned structured differences between the two persona-conditioned preference distributions. Therefore, the downstream experiment provides additional evidence that the proposed drift and direction labels are meaningful, and that the dataset supports persona-sensitive preference learning. Since VPL is designed to learn separable latent representations for divergent user preferences, these results strengthen the validity of our dataset for pluralistic and adaptive preference modelling.

\section{Prompts used for dataset generation}

\label{app:Dataset additional details}

The dataset was generated using a structured prompting strategy designed to produce multi-turn conversations exhibiting controlled stylistic tone drift while preserving semantic intent. For each communication domain, a dedicated prompt template was used to guide the generation process. All prompts enforce a fixed conversation length of six prompt–response pairs and require the assistant to rewrite the full response at every turn based on incremental user feedback.

Across all domains, the first prompt in each conversation requests an initial draft and is explicitly constrained to have no tone drift. Subsequent prompts introduce natural, user-like revision requests. Tone drift is annotated at the transition level and is allowed only when a noticeable change in tone occurs, such as formal to casual or vice versa. To ensure realism, each conversation contains a minimum of two and a maximum of three tone-drift transitions distributed organically across turns two to six.

All prompts enforce strict output constraints, including a fixed JSON schema, full response rewrites at every turn, and prohibition of placeholders. This design ensures consistency across domains while allowing stylistically diverse refinements. The following sections provide the exact prompt templates used for each communication genre.

\label{app:prompts}

\begin{center}
\setlength{\fboxsep}{6pt}
\setlength{\fboxrule}{0.5pt}

\fbox{%
    \includegraphics[width=1.0\linewidth]{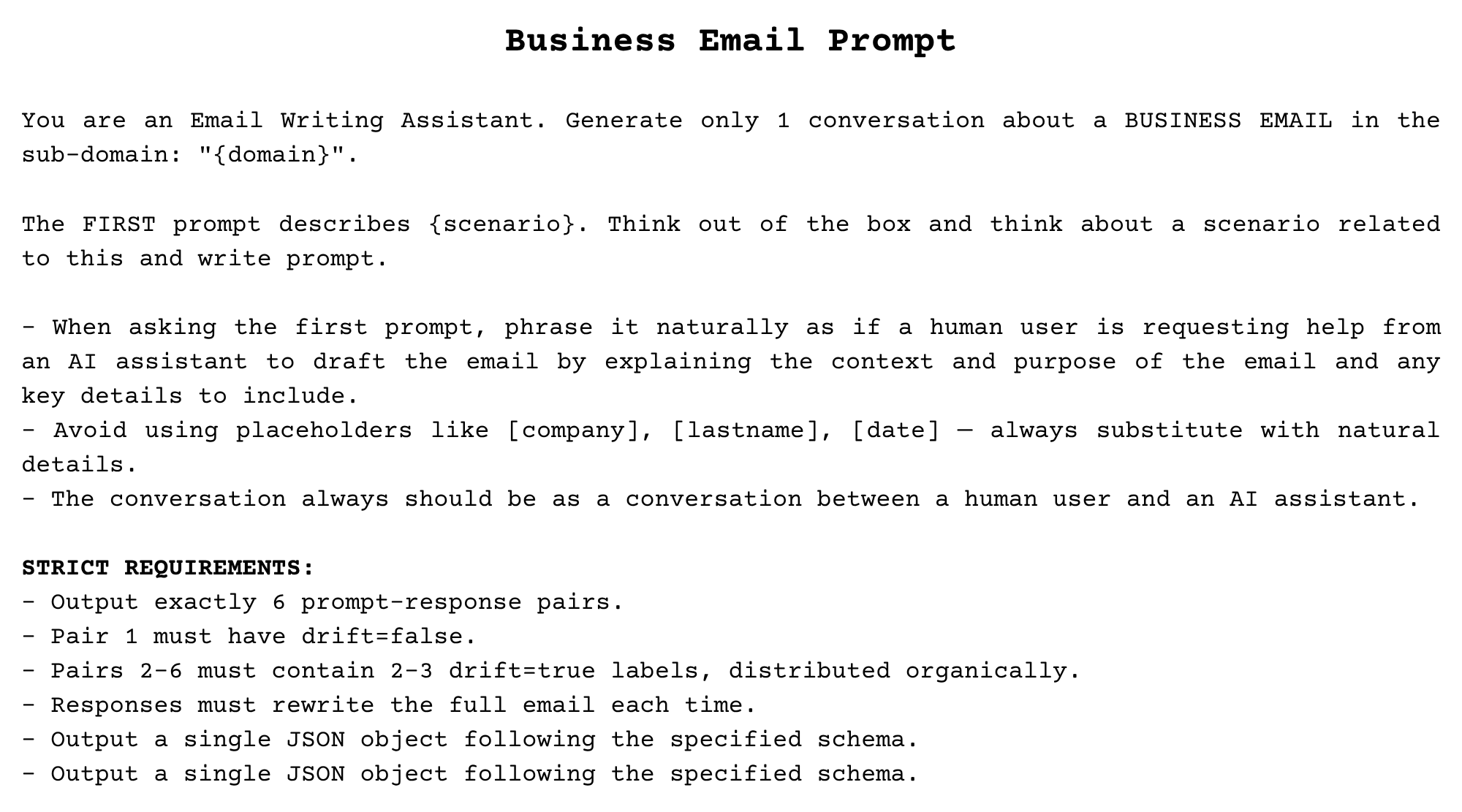}%
}

\captionof{figure}{Business Email prompt template used for dataset generation.}
\label{fig:business_prompt}
\end{center}

\vspace{0.3cm}

\begin{center}
\setlength{\fboxsep}{6pt}
\setlength{\fboxrule}{0.5pt}

\fbox{%
    \includegraphics[width=1.0\linewidth]{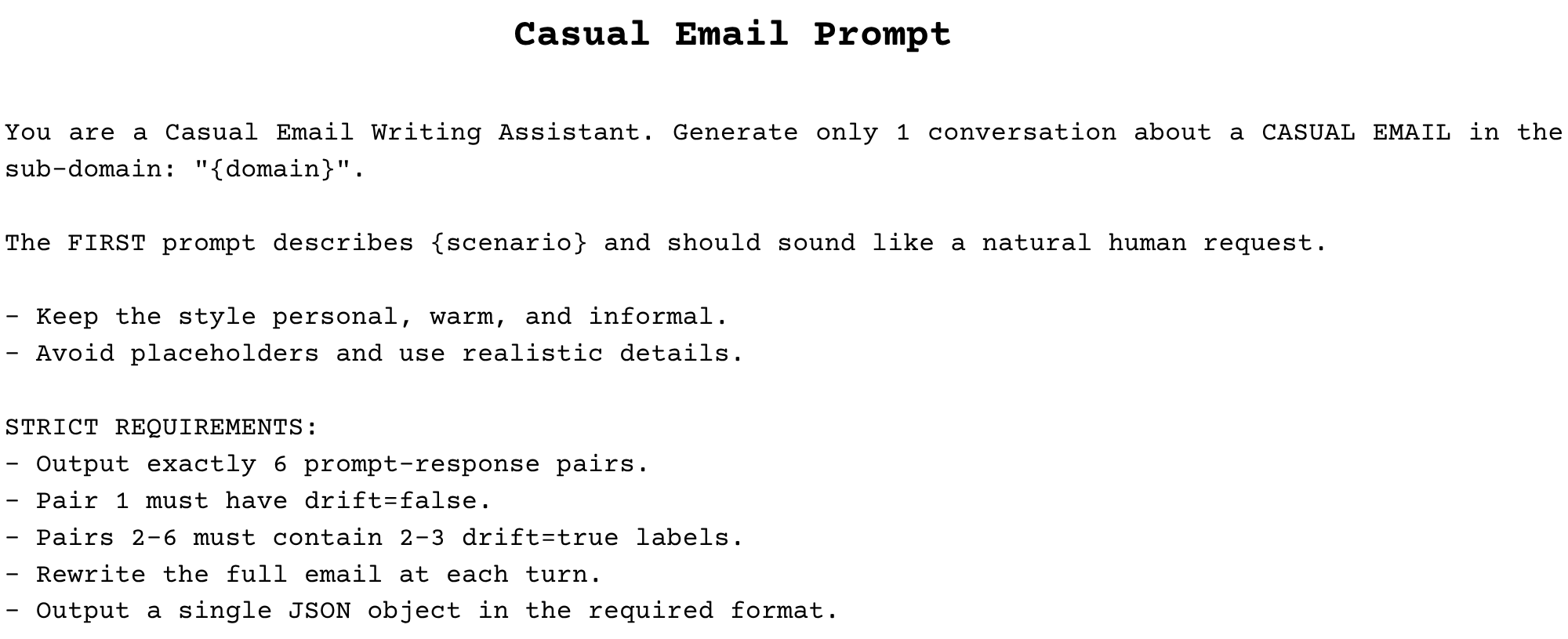}%
}

\captionof{figure}{Casual Email prompt template used for dataset generation.}
\label{fig:casual_prompt}
\end{center}

\vspace{0.3cm}

\begin{center}
\setlength{\fboxsep}{6pt}
\setlength{\fboxrule}{0.5pt}

\fbox{%
    \includegraphics[width=1.0\linewidth]{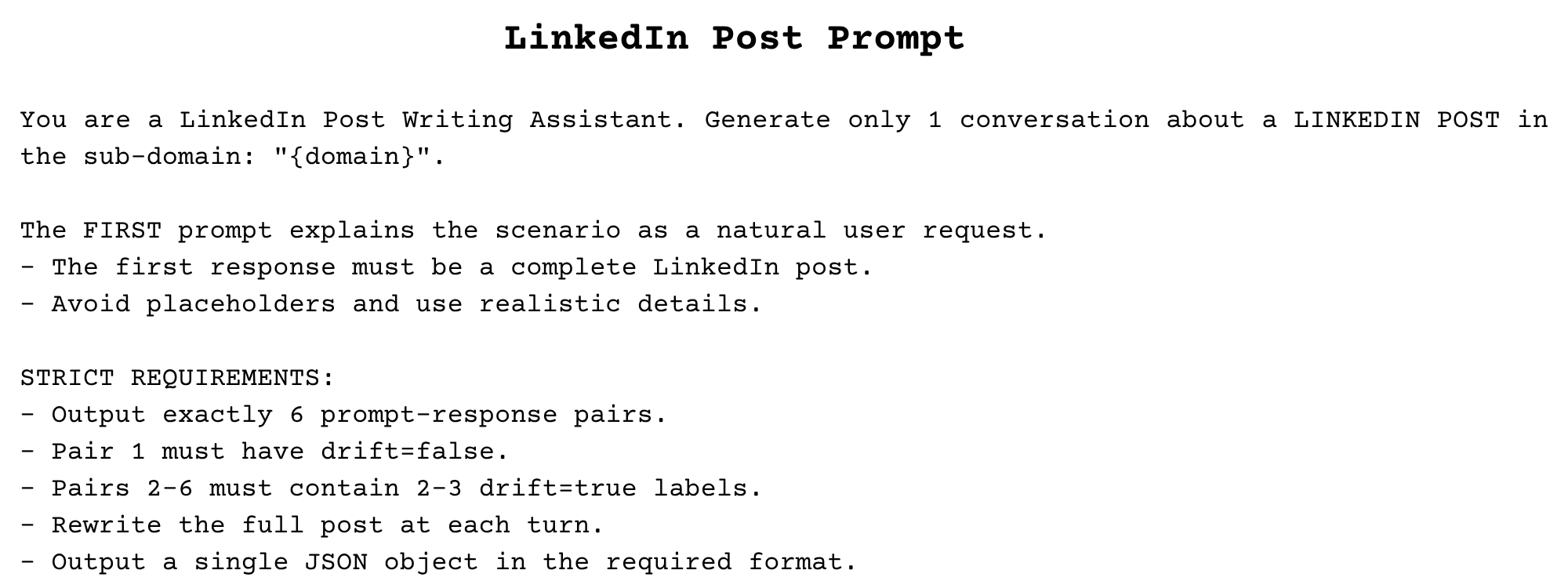}%
}

\captionof{figure}{LinkedIn post prompt template used for dataset generation.}
\label{fig:linkedin_prompt}
\end{center}

\vspace{0.3cm}

\begin{center}
\setlength{\fboxsep}{6pt}
\setlength{\fboxrule}{0.5pt}

\fbox{%
    \includegraphics[width=1.0\linewidth]{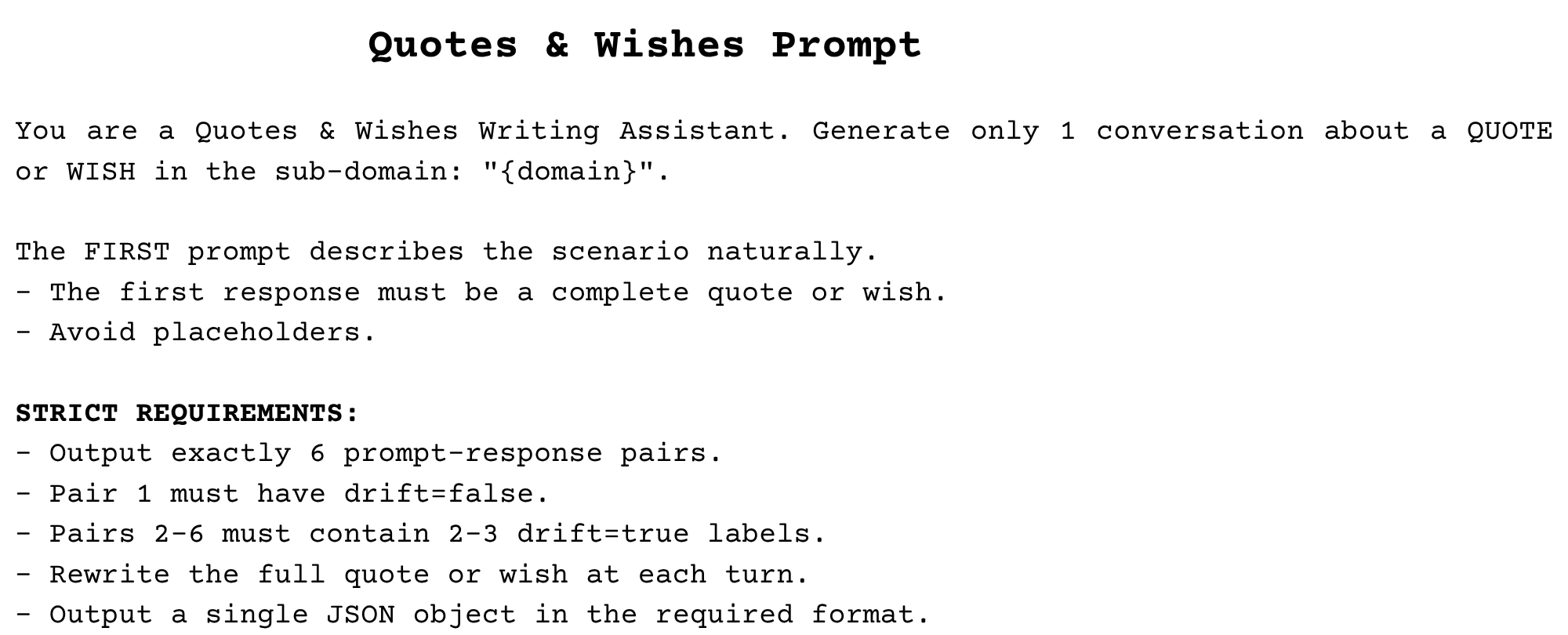}%
}

\captionof{figure}{Quotes and wishes prompt template used for dataset generation.}
\label{fig:casual_prompt}
\end{center}

\vspace{0.3cm}

\begin{center}
\setlength{\fboxsep}{6pt}
\setlength{\fboxrule}{0.5pt}

\fbox{%
    \includegraphics[width=0.9\linewidth]{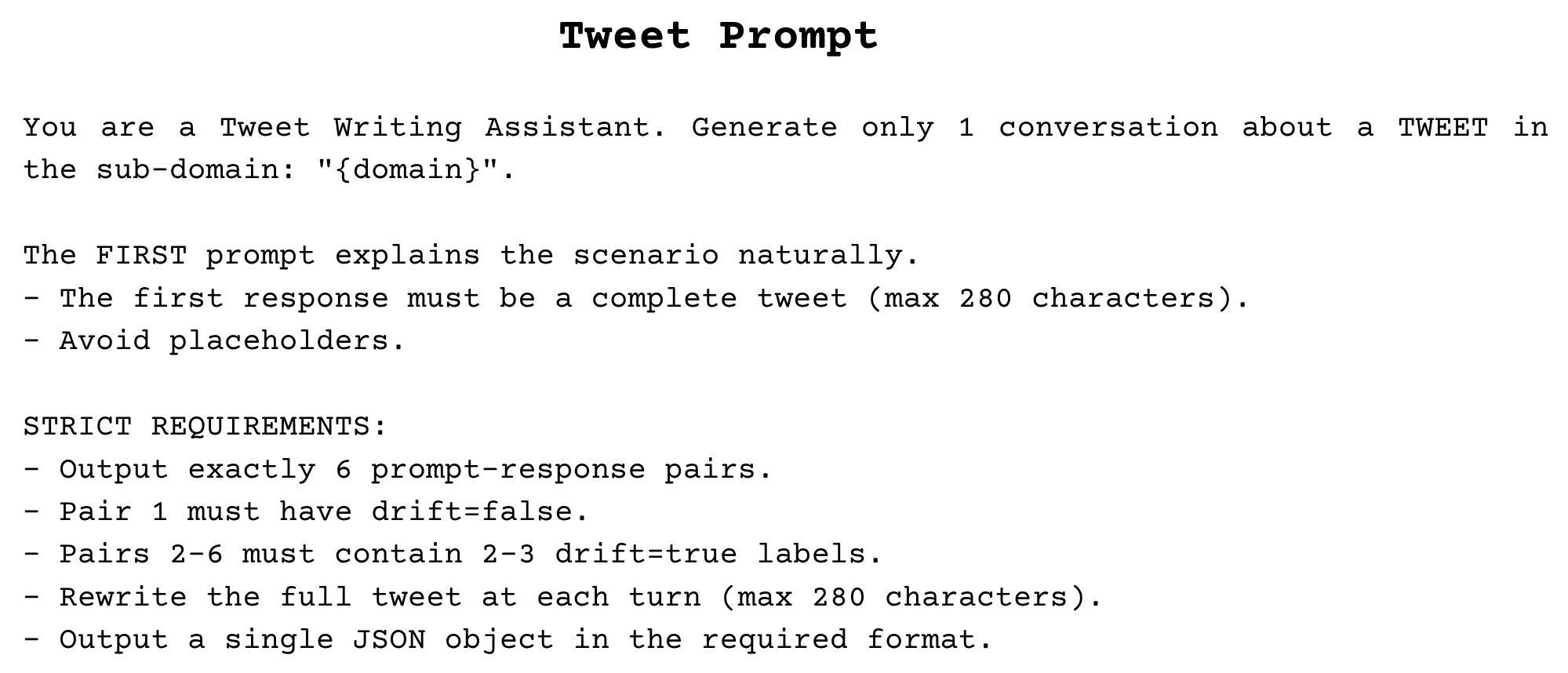}%
}
\captionof{figure}{Tweet post prompt template used for dataset generation.}
\label{fig:tweet_prompt}
\end{center}

\begin{figure}[H]
    \centering
    \setlength{\fboxsep}{6pt}   
    \setlength{\fboxrule}{0.5pt} 
    \fbox{
        \includegraphics[width=0.9\linewidth]{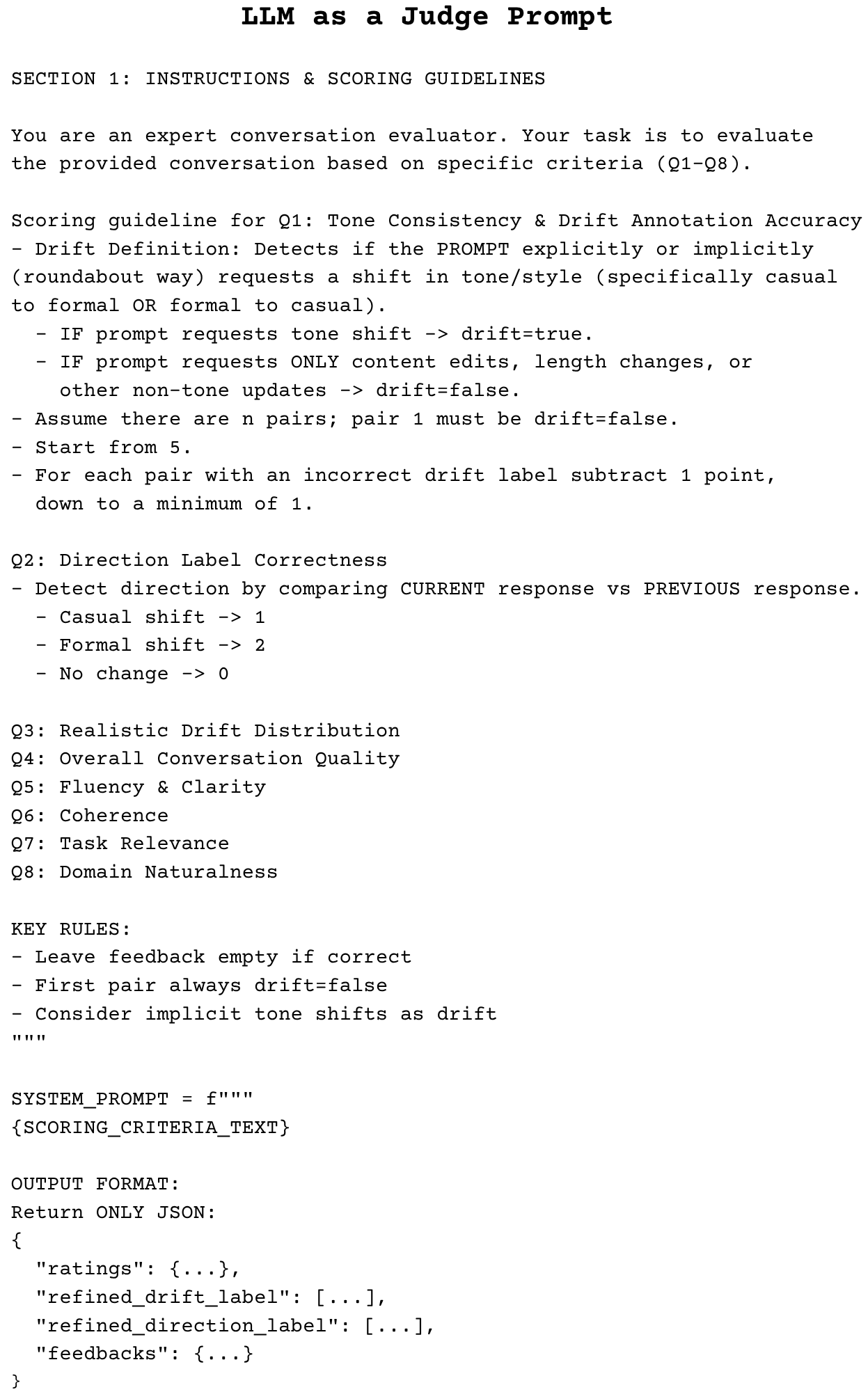}
    }
    \caption{Prompt template used for LLM as a Judge.}
    \label{fig:Judge_prompt}
\end{figure}




\section{Pairwise Inter-Persona Agreement}
\label{app:pairwise_agreement}
Table~\ref{tab:cohen_kappa_personas} presents the complete pairwise
$\kappa$ agreement scores underlying the persona-level distributions shown
in Figure~\ref{fig:persona_agreement_violin}. Positive values indicate
similar stylistic preferences between persona pairs, whereas negative values
indicate consistent preferences for opposing stylistic choices.

\begin{table}[H]
\centering
\resizebox{0.8\linewidth}{!}{
\begin{tabular}{lccccc}
\hline
 & A & B & C & D & E \\
\hline
A & 1.00 & -0.26 & 0.49 & 0.51 & -0.29 \\
B & -0.26 & 1.00 & -0.36 & -0.29 & 0.38 \\
C & 0.49 & -0.36 & 1.00 & 0.34 & -0.24 \\
D & 0.51 & -0.29 & 0.34 & 1.00 & -0.66 \\
E & -0.29 & 0.38 & -0.24 & -0.66 & 1.00 \\
\hline
\end{tabular}}
\caption{Pairwise Cohen's $\kappa$ agreement between personas.}
\label{tab:cohen_kappa_personas}
\end{table}

\section{Dataset Distribution Statistics}
\label{app:statistics}

\begin{figure}[H]
    \centering
    \includegraphics[width=0.9\linewidth]
    {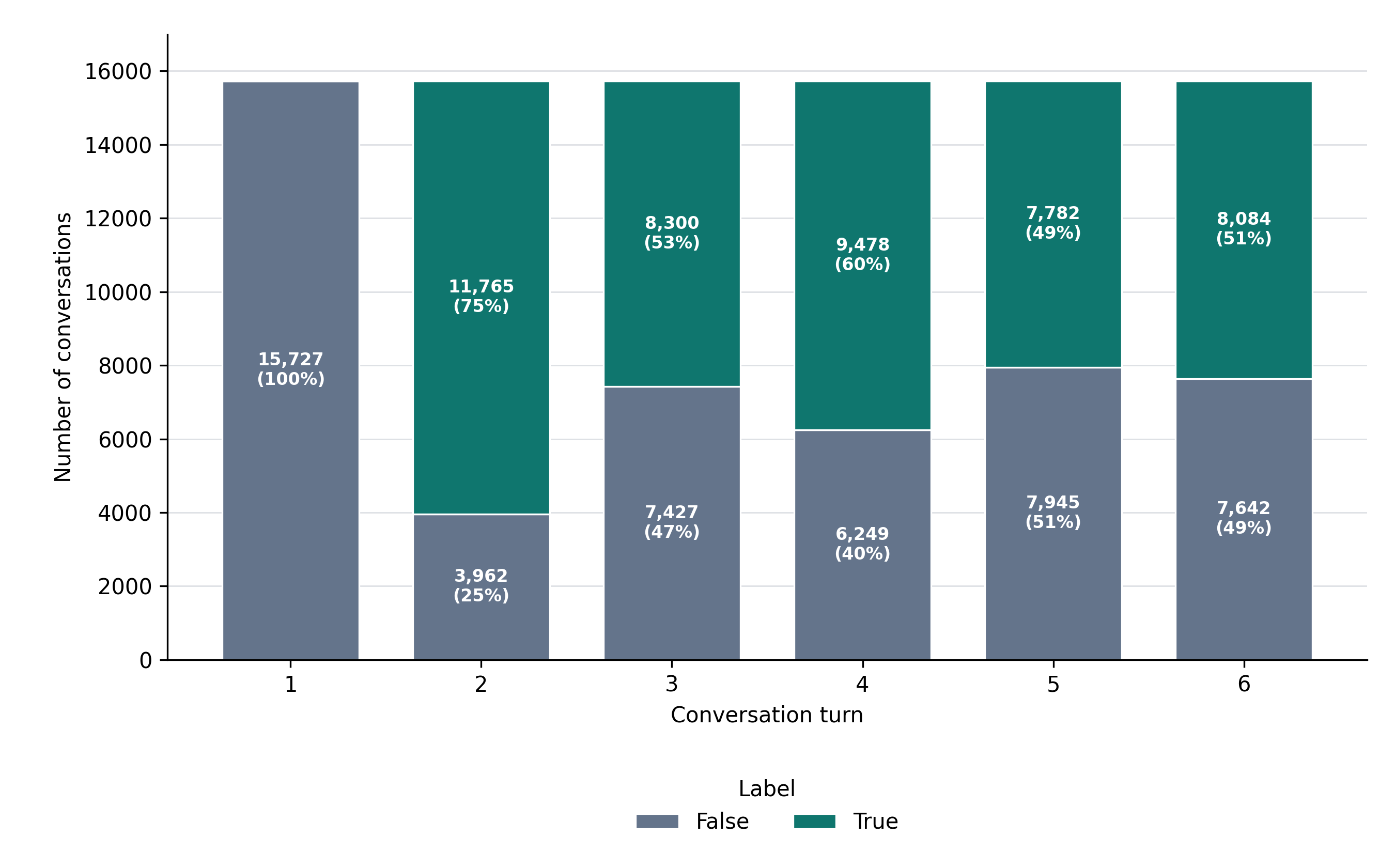}
    \caption{Distribution of drift labels across conversation turns.}
    \label{fig:drift_distribution}
\end{figure}

Figure~\ref{fig:drift_distribution} shows how drift changes throughout
the conversations. All first turns are labelled as non-drift because
they establish the initial conversational context. Drift is most common
in the second turn, occurring in 74.8\% of conversations. The proportion
decreases to 52.8\% in turn three, increases to 60.3\% in turn four, and
remains approximately balanced between drift and non-drift in turns five
and six.

\begin{figure}[H]
    \centering
    \includegraphics[width=0.9\linewidth]
    {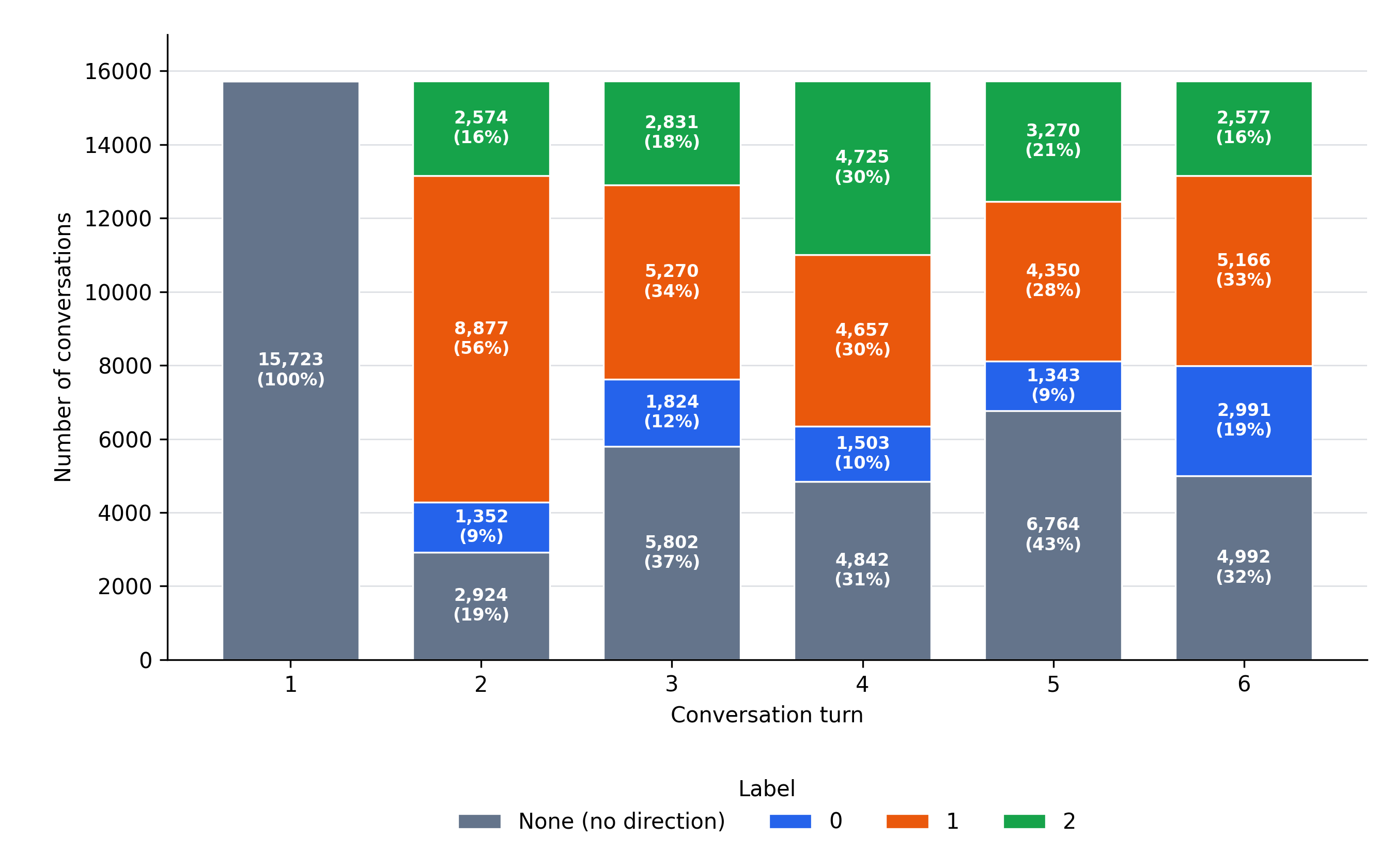}
    \caption{Distribution of direction labels across conversation turns.}
    \label{fig:direction_distribution}
\end{figure}

Figure~\ref{fig:direction_distribution} presents the distribution of
direction labels. The first turn predominantly has no direction because
it does not contain drift. Direction 1 is most frequent in turn two,
accounting for 56.4\% of conversations. The direction distribution
becomes more balanced in later turns, particularly in turn four, where
Directions 1 and 2 each represent approximately 30\% of conversations.

\begin{figure}[H]
    \centering
    \includegraphics[width=0.9\linewidth]
    {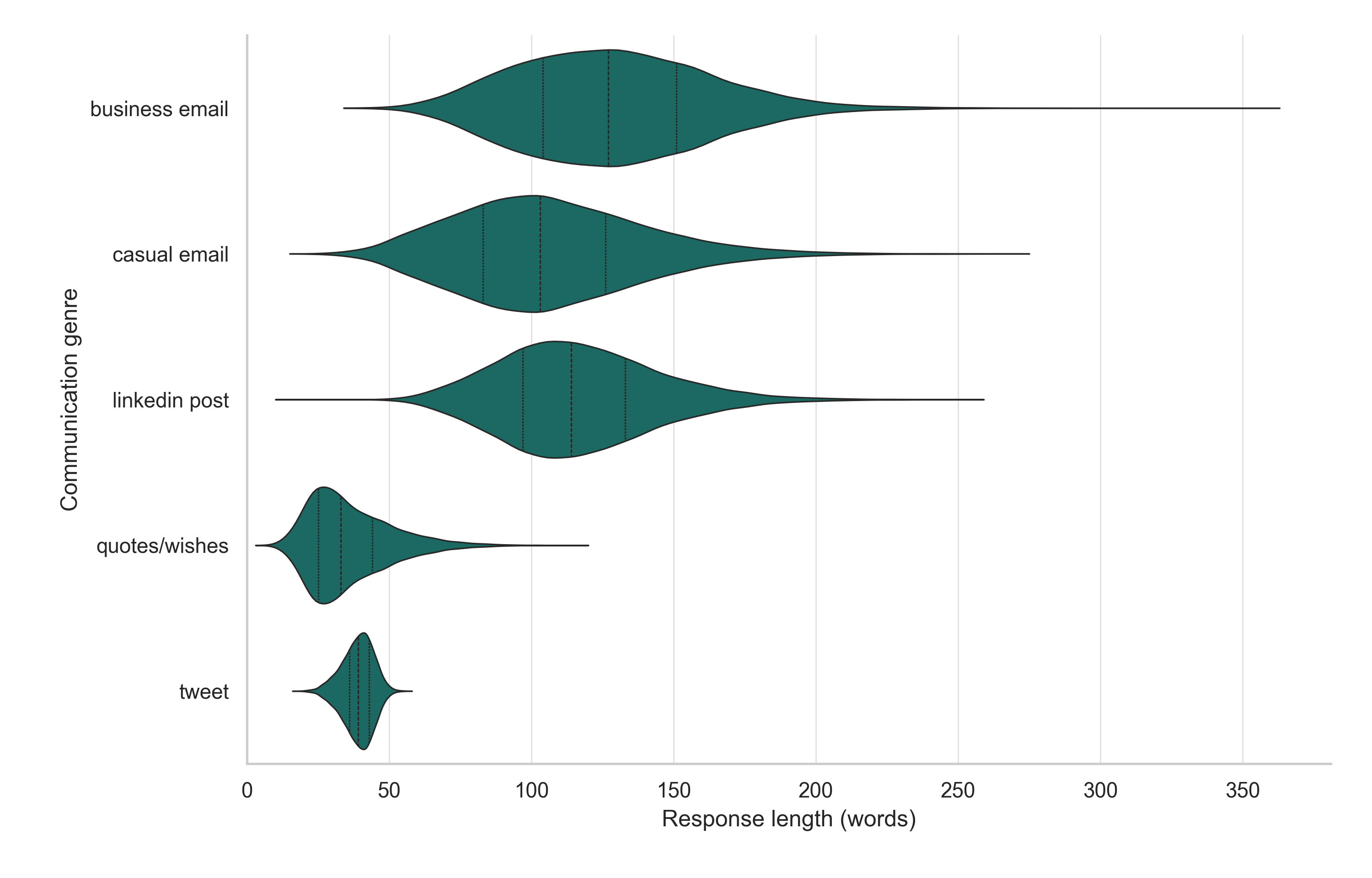}
    \caption{Distribution of response length by communication genre.}
    \label{fig:response_length_distribution}
\end{figure}

Figure~\ref{fig:response_length_distribution} shows response length,
measured in words, for each communication genre. Business emails have
the longest responses, with a median of 127 words, followed by LinkedIn
posts with 114 words and casual emails with 103 words. Quotes and wishes
and tweets are substantially shorter, with median lengths of 33 and
39 words, respectively. These differences are consistent with the
expected structural and stylistic requirements of each communication
genre.
\end{document}